\documentclass[runningheads,a4paper]{llncs}

\usepackage{amssymb}
\usepackage{graphicx}
\usepackage{amsmath}
\DeclareMathOperator*{\argmin}{\arg\!\min}

\usepackage{url}

\begin{document}

\mainmatter  

\title{Kernel classification of connectomes based on earth mover's distance between graph spectra}

\titlerunning{Kernel classification of connectomes based on EMD between spectra}

%
%
\author{Yulia Dodonova$^{1, 2}$
\and Mikhail Belyaev$^{2, 1}$\and Anna Tkachev$^{1}$\and \\Dmitry Petrov$^{1, 2}$\and Leonid Zhukov$^{3}$}
\authorrunning{Dodonova, Belyaev, Tkachev, Petrov, Zhukov}

\institute{1 - Kharkevich Institute for Information Transmission Problems, Moscow, Russia \\
2 - Skolkovo Institute of Science and Technology (Skoltech), Moscow, Russia\\
3 - National Research University Higher School of Economics, Moscow, Russia\\
ya.dodonova@mail.ru, belyaevmichel@gmail.com, annatkachev42@gmail.com,\\ 
to.dmitry.petrov@gmail.com, lzhukov@hse.ru
}

%
%

\toctitle{Lecture Notes in Computer Science}
\tocauthor{Authors' Instructions}
\maketitle

\begin{abstract}
In this paper, we tackle a problem of predicting phenotypes from structural connectomes. We propose that normalized Laplacian spectra can capture structural properties of brain networks, and hence graph spectral distributions are useful for a task of connectome-based classification. We introduce a kernel that is based on earth mover's distance (EMD) between spectral distributions of brain networks. We access performance of an SVM classifier with the proposed kernel for a task of classification of autism spectrum disorder versus typical development based on a publicly available dataset. Classification quality (area under the ROC-curve) obtained with the EMD-based kernel on spectral distributions is 0.71, which is higher than that based on simpler graph embedding methods. 
\end{abstract}

\section{Introduction}

Machine learning prediction of brain disorders based on neuroimaging data has gained increasing attention in recent years. Studies aiming at predicting brain disorders based on voxel-level and region-level MRI features are reviewed in \cite{ml_review} and \cite{Plis}; these reviews also highlight some important pitfalls of machine learning based analyses of neuroimaging data.

When it comes to classification of connectomes, the task becomes even more challenging. Mathematically, the input objects of machine learning algorithms are now graphs with particular properties. Let $X_i$ be a macroscale brain network, $y_i$ be a class label (we only deal with $y_i\in\{0,1\}$ throughout this study). Given a training set of pairs $(X_i, y_i)$ and the test set of input objects $X_j$, the task is to make a good prediction of the unknown class label $y_j$. The task is specific because of the particular objects $X_i$: these are relatively small undirected fully connected graphs with weighted edges and uniquely labeled nodes which are localized in 3D space. For structural connectomes, edges are weighted proportionally to the number of streamlines between the brain regions.

The most obvious approach to classification within these settings would be to adopt some strategy of transforming adjacency matrices into vectors from $\mathbb{R}^p$ because most classifiers work with this type of input objects. In this paper, we develop a different approach. 

We propose that spectra of the normalized graph Laplacians can capture the most essential structural properties of brain networks. Hence, spectral distributions of brain networks can differ in normal and pathological development. To account for differences in the spectral distributions of connectomes we propose to use an earth mover's distance (EMD), a metric that appears in transportation problem as a discretized version of the mass transportation distance. We next use the pairwise distances between spectral distributions of brain networks to construct a kernel for a support vector machines (SVM) classifier.
We evaluate the proposed approach for a task of classification of autism spectrum disorder versus typical development based on a publicly available dataset. We report classification quality that is higher than that obtained based on simpler graph embedding methods.

\section{Spectral representation of graphs}
We propose that spectra of the normalized Laplacians can be used to meaningfully represent connectomes. Given an undirected weighted graph with $n$ nodes, we define the adjacency matrix $A$ as the  $n\times n$ matrix with entries $a_{ij}$, where $a_{ij}$ is the weight between the respective nodes (the weights are not necessarily binary). 
We define $D$ as the diagonal matrix of weighted node degrees: 

\begin{equation}
d_i = \sum_{j}a_{ij}.
\end{equation}

The normalized graph Laplacian is given by:
\begin{equation}
\mathcal{L} = D^{-1/2}(D - A)D^{-1/2}
\end{equation}



\begin{figure}
\centering
\includegraphics[scale=0.52]{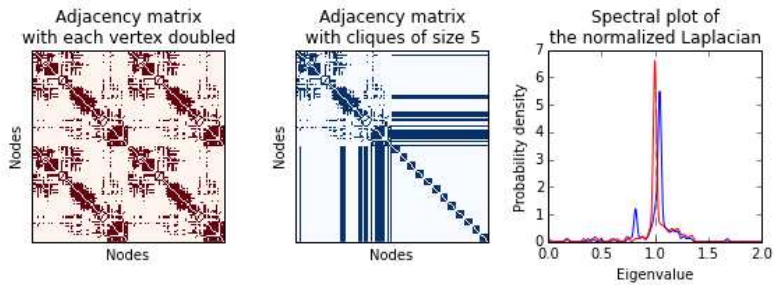}
\caption{Some examples of how particular graph structures appear in the distributions of the normalized Laplacian spectra a graph produced from the same original matrix by doubling each vertex (left) and adding identical cliques of size 5 (middle), and  the distributions of the normalized Laplacian eigenvalues of these two graphs (right). A fragment of a real unweighted connectome is used as the original matrix; Gaussian kernel density reconstruction is used to plot distributions.}
\label{fig001:example}
\end{figure}

For an extensive theory on the normalized Laplacian spectra, we refer to \cite{Chung}. The spectra of the normalized Laplacian are particularly informative because they have some useful properties. The eigenvalues are always in range from 0 to 2; this allows the comparison of networks with different sizes. Importantly, the overall shape of the eigenvalue distribution, its symmetry and the multiplicity of particular values also capture an important information about graph structure; Figure 1 gives an example illustration on a toy network. For the formal definitions and proofs, we refer to \cite{Banerjee-2008a}.

A paper \cite{Banerjee-2008b} also plots many examples of the spectral distributions of the normalized Laplacians of the simulated and real networks; the latter include biological networks, weblog hyperlink graphs, power grids, US football games, and many others. The authors attempt to group these networks into classes based on the shape of their spectral distributions; they suggest that the shape of the spectral distribution of the networks of a particular class can provide insights into their structure and growth. However, the proposed classification scheme is rough and is based only on the visual evaluation of the plotted distributions.

A similar work is done in \cite{de-Lange}. The authors compare eigenvalue distributions of the normalized Laplacians across the structural brain networks of the cat, macaque and Caenorhabditis elegans; they describe similarities in these networks and also compare them to some random networks and real networks from other domains.  Based on the visual evaluation of the plotted distributions, the authors claim that the close relation between the normalized Laplacian spectra of the analyzed networks suggests shared underlying structural properties of the neuronal systems of the macaque, cat and
C. elegans. They conclude that the analysis of graph spectra reveals the structural brain networks to be in a specific class of networks, distinct from the other network classes.

Taken together, these papers suggest that the graph spectra can serve as informative features for the task of brain network classification. Hence, the problem of classifying graphs transforms into a problem of classifying distributions of the eigenvalues associated with these graphs. In the next section, we discuss how this can be accommodated.

\section{Classification of graphs based on distance between spectral distributions}

As discussed above, we aim to take into account the shape of the spectral distributions and the multiplicity of particular eigenvalues rather than simply consider the eigenvalues as feature vectors. Hence, we need to introduce a distance between spectral distributions. Provided  that we can measure the pairwise differences between the distributions of graph spectra, we can construct a kernel based on these differences. Afterwards, we can use the kernel trick and feed the obtained kernel to an SVM classifier.

\subsection{SVM classifier based on a distance matrix}
The SVM classifier is able to accept any input objects, not necessarily a set of vectors from $\mathbb{R}^p$. This means that any positive semi-definite function 

\noindent $K(\mathbf{x}_i, \mathbf{x}_j):~\mathbb{X}^2~\to~\mathbb{R}$ on the input data $\mathbb{X}$ can be used as a kernel for the SVM classifier provided that:
$$
\sum_{i=1}^N\sum_{j=1}^N K(\mathbf{x}_i, \mathbf{x}_j) c_i c_j \geq 0
$$
for any $(x_1, x_2, \dots, x_N) \in \mathbb{X}$ and any coefficients $(c_1, c_2, \dots, c_N) \in \mathbb{R}$. There are no constraints on the structure of the input data $\mathbb{X}$. Examples of kernels on graphs can be found in \cite{Gartner} and \cite{thesis}. 

In this study, we explore behavior of a kernel that is based on distances between graph spectral distributions.
Let $\mathcal{S}_i$ be the spectrum of a normalized Laplaсian $\mathcal{L}_i$ and $\Delta(\mathcal{S}_i, \mathcal{S}_j)$ be a distance between spectra $\mathcal{S}_i, \mathcal{S}_j$.  We build a graph kernel $K$ using the distance $\delta$ as follows:
\begin{equation} \label{eq:kernel}
K(\mathcal{S}_i, \mathcal{S}_j) = e^{- \delta(\mathcal{S}_i, \mathcal{S}_j)}
\end{equation}

\subsection{Earth mover's distance between spectral distributions}

Hence, we need to compute pairwise distances between spectral distributions. There are two possible ways to accommodate this. First, we can estimate the density using empirical values and use measures that work with localized data representations. We explored this method in our previous work \cite{mlsp}. We first split the entire range of eigenvalues into equal intervals (bins) and computed frequencies within each bin.  We next considered the two probabilistic distance measures: the Kullback-Leibler divergence and the Jensen-Shannon distance. We obtained the information divergence based kernels by exponentiating these measures and examined their performance in the tasks of classification of autism spectrum disorder versus typical development and of the carriers versus non-carriers of an allele associated with the high risk of Alzheimer disease.
The outcomes of the algorithm were extremely sensitive to the number of bins used to reconstruct density. On average, the classification quality was rather poor, with the area under the receiver operating characteristic curve (ROC AUC) of about 0.65, although with some arbitrary numbers of bins it occasionally unsystematically peaked higher. 

In this study, we introduce a different approach that does not require defining arbitrary bin sizes and hence overcomes the major weakness of the previously used method.
The algorithm proposed here compares the spectral distributions directly based on the vectors of eigenvalues. 
To accommodate this, we propose to use the earth mover's distance (EMD) \cite{EMD}, which measures the minimum cost of transforming one sample distribution into another.  
This metric receives its name from the idea behind it: if each distribution is represented by some amount of dirt, EMD is the minimum cost required to move the dirt of one distribution to produce the 
other. The cost is the amount of dirt moved times the distance by which it is moved.

The EMD is thus based on the solution to the transportation problem, which is a problem of finding a least expensive flow of goods from the suppliers to the consumers that satisfies the consumers' demand. From this point, the EMD can also be viewed as a discretized version of the Monge-Kantorovich mass transportation distance, which is defined for continuous measures. In statistics, the EMD is also known as the Wasserstein metric between probability distributions. We refer to \cite{Vershik} for a review on the history of this measure.

More technically, let $\{s^i_1, ..., s^i_n\}$ be the eigenvalues of spectrum $\mathcal{S}_i$. We put an equal measure $1/n$ to each point $s^i_k$  on real line. Let $f_{kl}$ be the flow of mass between the points $s^i_k$ and $s^j_l$. The EMD is the normalized flow of mass between sets $\mathcal{S}_i = \{s^i_1, ..., s^i_n\}$ and $\mathcal{S}_j = \{s^j_1, ..., s^j_n\}$ that minimizes the overall cost:
\begin{equation}\label{eq:emd}
emd(\mathcal{S}_i, \mathcal{S}_j) = 
\argmin_{F=\{f_{kl}\}} \frac{\sum_{k,l} f_{kl} |s^i_k - s^j_l|}{\sum_{k,l}f_{kl}},
\end{equation}
with the constraints: 
$f_{kl}\geq 0$, \qquad$\sum_{k=1}^n f_{kl} =  1/n$, \qquad
$\sum_{l=1}^n f_{kl} = 1/n$.

The EMD is a metric, and hence the kernel \eqref{eq:kernel} is a true positive semi-definite kernel. In what follows, we examine how SVM classifier with this kernel performs in a task of classification of autism spectrum disorder versus typical development.

\section{Data}
We explore the performance of the proposed kernel using a publicly available UCLA Autism dataset (described in \cite{Rudie-2013}). This dataset includes DTI-based connectivity matrices of 51 high-functioning autism spectrum disorder (ASD) subjects (6 females) and 43 typically developing (TD) subjects (7 females). Average age (standard deviation) is 13.0 (2.8) for ASD group and 13.1 (2.4) for TD group. 

Nodes are defined using parcellation scheme that is based on a large meta-analysis of functional MRI studies combined with whole-brain functional connectivity mapping. This scheme produces 264 equal-size brain regions and thus 264$\times$264 connectivity matrices. Edges result from brain deterministic tractography. It is performed on voxelwise fractional anisotropy values using the fiber assignment by continuous tracking (FACT) algorithm \cite{FACT}. Edge weights in the original matrices are proportional to the number of streamlines detected by the algorithm.

\section{Methods}

\subsection{Network construction}

For each dataset, we apply three different weighting schemes. Based on our previous findings\cite{prni}, we expect that connectomes that capture information on both strength (number of streamlines) and lengths (physical distances between brain regions) of the connections  are most useful for classification, at least for this particular task. For a purpose of comparison, we also consider weightings that only take into account either strength or distance. For the former, we take the original weighted connectivity matrices. For the latter, we first binarize original connectomes and than scale the existing edges by their lengths:
\begin{equation}\label{eq:len}
a^{dist}_{ij} = \frac{1 \;\; if \;\; a_{ij} > 0 \;\;  else \;\; 0}{l_{ij}},
\end{equation}
where $l_{ij}$ is the Euclidean distance between centers of the regions $i$~and $j$. We use Montreal Neurological Institute (MNI) coordinates of region centers provided by the authors of the datasets to obtain the reasonable proxy of the distances between brain regions.
Third, we combine the two weighting schemes and produce connectomes with weights defined by:
\begin{equation}\label{eq:comb}
a^{combined}_{ij} = \frac{a_{ij}}{l_{ij}},
\end{equation}

For each of the three weighting schemes, we next scale each matrix by the sum of its edge weights: $
a^{scaled}_{ij} = \frac{a_{ij}}{\sum_{i, j} a_{ij}}$.
This scaling does not produce any changes in the spectra of the normalized Laplacians; hence, this step is not needed for our proposed pipeline. However, we also use "bag of edges" as the baseline features (to be described in the next section), and we use scaling to enhance between-subject comparison based on these features.

\subsection{Classification pipeline}
For each of the three weighting schemes, we produce the normalized Laplacians of the connectomes and compute their spectra. We next compute the EMD-based kernels by \eqref{eq:kernel}, \eqref{eq:emd}. To compute the pairwise EMD between spectra we use an implementation that uses the original samples of points and does not require any data reduction. 
The kernels are then fed to the SVM. The regularization parameter of the SVM classifier is fixed at 0.1, 1, 10, and 50. We report the results obtained for the models with the best value of this penalty parameter.

We use two different approaches to produce baseline classification quality. First, we use the same eigenvalues of the normalized Laplacians and treat them as feature vectors. We run linear kernel SVM on the vectors of eigenvalues. The regularization parameter of the SVM classifier is fixed at the same values as above.
Second, we vectorize each weighted adjacency matrix by taking the values of its upper triangle (the so-called "bag of edges") and use these vectors as baseline features. Again, we run linear kernel SVM on these features with the penalty parameter fixed at the same values as above.

We use the ROC AUC to assess the predictive quality of the algorithms. We use 10-fold cross-validation and combine predictions on all test folds to estimate the classification quality on the entire sample. We repeat this procedure 100 times with different 10-fold splits, thus producing 100 ROC AUC values. 

We use Python 2.7 and IPython notebooks platform, specifically NumPy, SciPy, pandas, matplotlib, scikit-learn \cite{skl}, and pyemd \cite{pyemd} libraries. All scripts will be available at https://github.com/YuliaD/BACON-2016.

\section{Results and discussion}
\subsection{Distributions of the normalized Laplacian eigenvalues}

We first plot spectral distributions of the connectomes in Figure 2. We use the group average ASD and TD matrices with the combined weights \eqref{eq:comb}, compute the associated $\mathcal{L}$ and its spectrum. To produce the plots, we convolve the eigenvalues with a Gaussian kernel; that is, we plot the function: $f(x) = \sum_{\lambda_j}\frac{1}{\sqrt{2\pi\sigma^2}}exp(-\frac{|x-s_j|^2}{2\sigma^2})$. 
We use $\sigma=0.02$. 
For purpose of illustration, we also produce three random graphs and plot their respective spectra in Figure 2. The first one is the Erd$\ddot{o}$s-R$\acute{e}$nyi (ER) graph for which the required number of connections is produced randomly. The second random graph is based on the Barab$\acute{a}$si-Albert (BA) preferential attachment model. The third graph is the Watts-Strogatz (WS) network with small-world properties (the rewiring probability is set to 0.2). The number of nodes and edges in these random graphs is set equal to that of the group average matrix of the real dataset. The plotting algorithm is the same as for the actual matrices.

The plots produced for the human brain connectivity matrices  differ from the respective plots obtained for the random graphs. Of the three random graph models, the small-world WS model produces a spectral distribution most close to that of the connectomes.

Empirical distributions based on human brain networks are close to those of the cat, macaque and C. elegans described by \cite{de-Lange}. Our plots exactly follow their description of the respective plots observed for the non-human connectomes: (1) the spectra are skewed to the left with the largest eigenvalue much closer to one than the smallest eigenvalue; (2) the distributions show a peak close to one; and (3) the smallest eigenvalues are scattered around a few small peaks at the beginning of the spectra. In general, this confirms an observation \cite{de-Lange} that spectral distributions of brain networks share similar properties.

However, average ASD and TD spectral distributions are very close and almost coincide. At the same time, the distributions of the normalized Laplacian eigenvalues of the individual connectomes shown in Figure 2 largely vary. 

\begin{figure}
\centering
\includegraphics[scale=0.38]{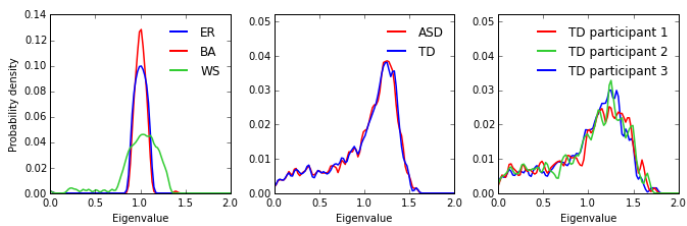}
\caption{Distributions of the normalized Laplacian eigenvalues of the random graphs (left), average real connectomes (middle) and example individual connectomes of typically developing subjects (right). Random graphs are unweighted; average and individual connectomes are weighted by \eqref{eq:comb}. Note different scales for the first plot.}
\label{fig01:example}
\end{figure}


\subsection{Classification based on spectral distributions}

Figure 3 shows the classification results of the SVM with the EMD-based kernel on graph spectra, the linear kernel on graph spectra, and the linear kernel on the "bag of edges".

Interestingly, the linear classifier that works with the "bag of edges" slightly outperforms a similar classifier that works with the vectors of eigenvalues. This is surprising due to a huge dimensionality of the vectors of edges (equal to the number of elements in the upper triangle of a connectivity matrix). This also shows that the eigenvalues of the normalized Laplacian spectra per se do not help to represent connectome objects in a space where they would be separable.

For the weighting schemes that are based solely on the strengths of the network connections or their physical lengths none of the algorithms works satisfactorily. However, for a weighting scheme that incorporates information on both strengths and lengths of the connections, the proposed pipeline works well and clearly  outperforms the baselines. The ROC AUC averaged over 100 runs of the algorithm is 0.706, standard deviation is 0.022. 

First, this partly confirms our previous finding that weighting connectomes by both the number  of streamlines and the physical distances enhances classification quality. This might be because long streamlines are more prone to bias of the tractography algorithm; downweighting long connections partly eliminates this bias. Second, the classification quality obtained with the EMD-based kernel SVM is better than the average classification results obtained in our previous work \cite{mlsp} using the probabilistic kernels. Importantly, the EMD-based kernel  is certainly

\begin{figure}
\centering
\includegraphics[scale=0.55]{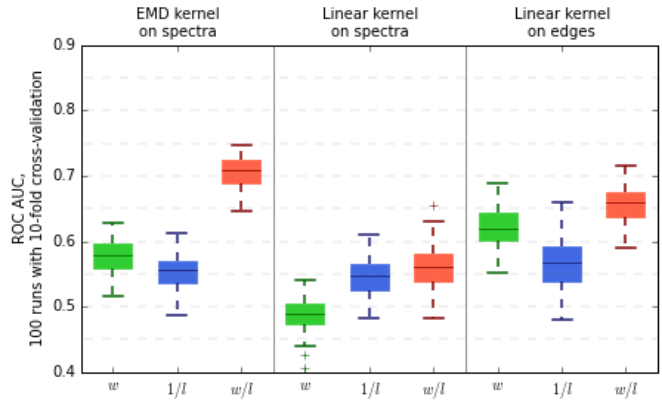}
\caption{Classification results obtained with the EMD kernel SVM on the eigenvalues of the normalized Laplacian, the linear kernel SVM on the eigenvalues of the normalized Laplacian, and the linear kernel SVM on the vectors of edges of the connectivity matrices. $w$, $1/l$ and $w/l$ refer to the matrices with the original weights and the weights obtained by \eqref{eq:len} and \eqref{eq:comb}, respectively. Each boxplot shows a distribution of ROC AUC values over 100 runs of the algorithm with different 10-fold splits.}
\label{fig1:example}
\end{figure}

\begin{figure}
\centering
\includegraphics[scale=0.445]{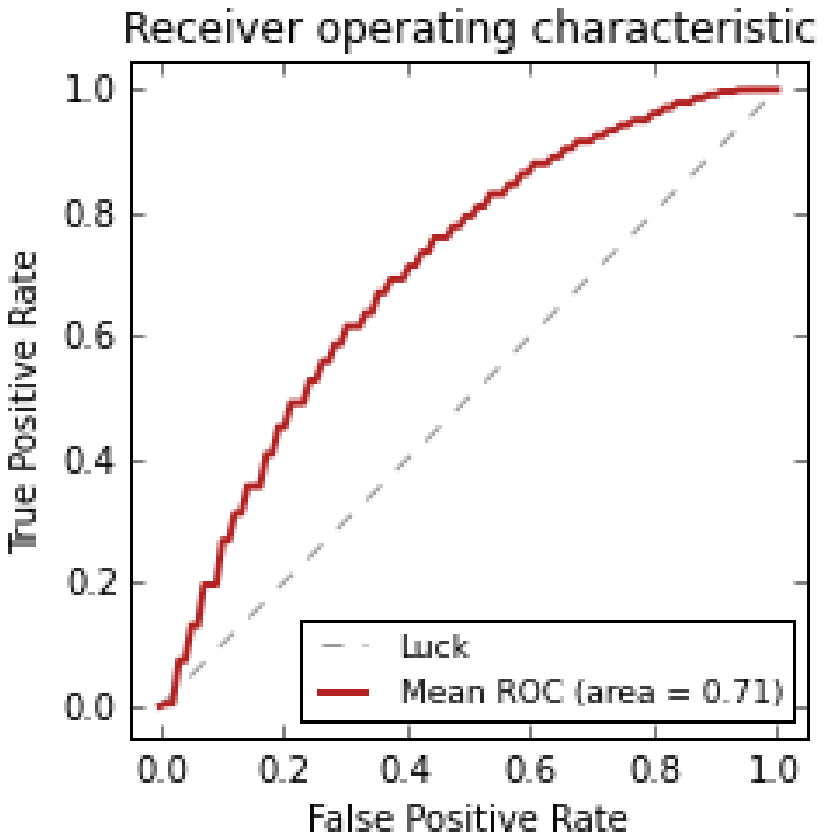}
\includegraphics[scale=0.42]{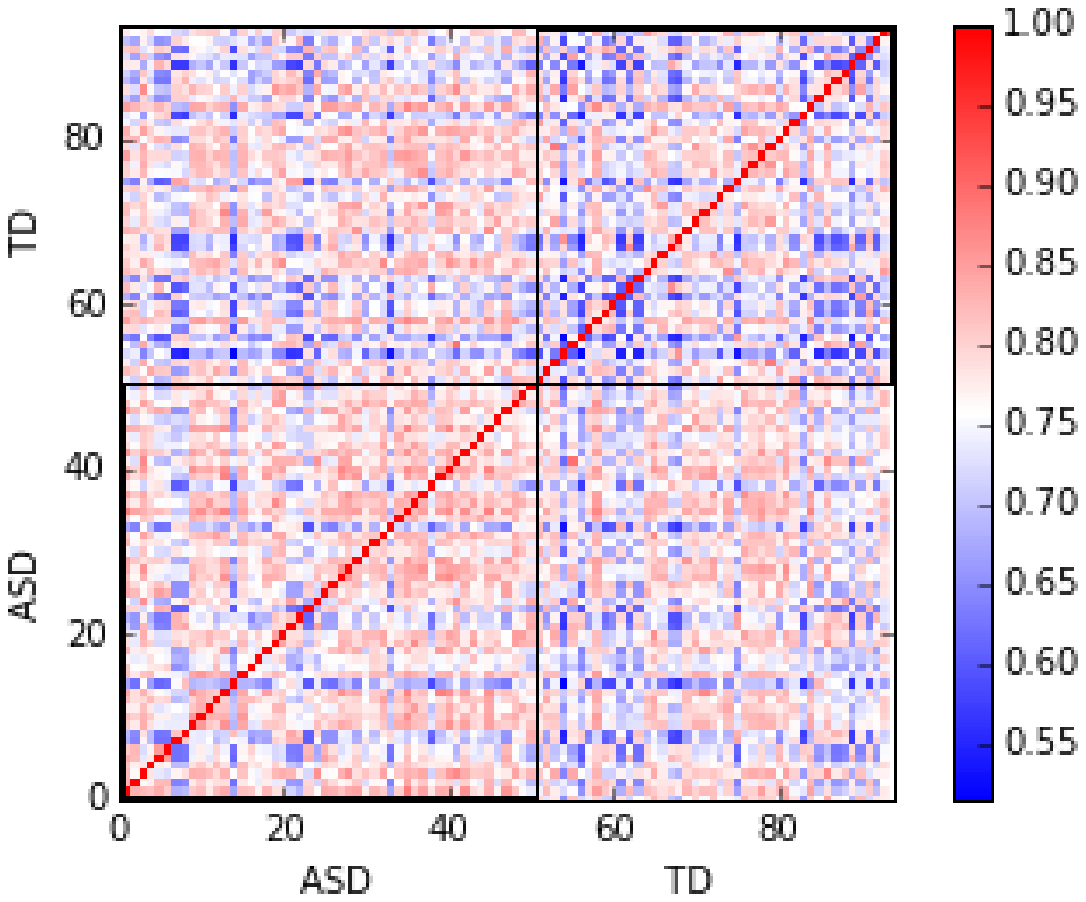}
\includegraphics[scale=0.505]{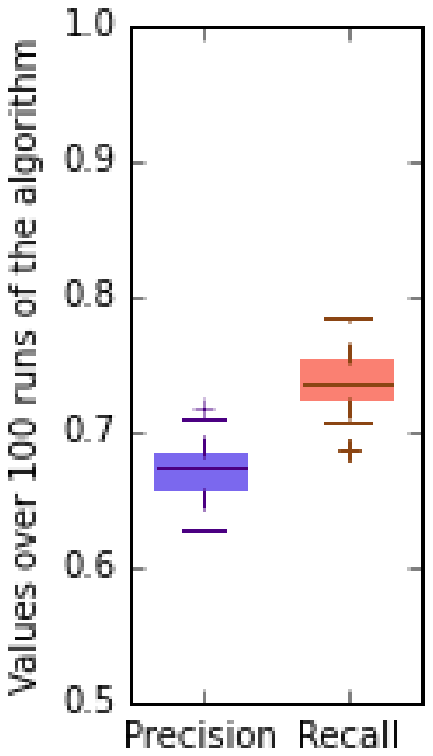}
\caption{Left: Average ROC curve obtained with the EMD kernel on the normalized Laplacian spectra; the mean value is the area under the interpolated average curve. Middle: Gram matrix based on the earth mover's distances between the normalized Laplacian spectra; the squares show submatrices for the ASD (lower left) and TD (upper right) participants. Right: Precision and recall values over 100 repetitions of the SVM with the EMD kernel on the normalized Laplacian spectra.}
\label{fig2:example}
\end{figure}

\noindent more robust in the sense that it does not require an intermediate step of density reconstruction but still allows to evaluate differences between the empirical eigenvalue distributions.

Figure 4 provides more details on how the EMD-based SVM classifier performed on our dataset. The ROC curve is shown in the left. The middle plot shows a Gram matrix obtained by \eqref{eq:kernel}, \eqref{eq:emd}; the submatrices for the ASD and TD subjects are in the lower left and upper right squares, respectively. This matrix suggests that the ASD subjects are somewhat closer to each other, while the TD group shows larger variability in pairwise difference values. In line with this, the right plot in Figure 4 shows that our algorithm performs quite well identifying ASD subjects (average recall value over 100 runs of the algorithm is 0.737, standard deviation is 0.025). At the same time, the algorithm somewhat tends to classify TD subjects as pathological (average precision value is 0.670, standard deviation is 0.019).

\section{Conclusions}
We propose an algorithm that uses eigenvalue distributions of the normalized Laplacian spectra of structural brain networks to classify normal and pathological development based on connectomes. We compute pairwise distances between spectral distributions of connectomes using earth mover's distance (EMD), a metric that appears as a solution to the transportation problem. Its main advantage over the previously explored probabilistic distances is that it takes into account the shapes of the spectral distributions and at the same time does not require density reconstruction from the empirical values. We compute a kernel based on this metric and run SVM classifier with this kernel. We examine the performance of the proposed algorithm for a task of classification of autism spectrum disorder versus typical development based on a publicly available dataset. The best-performing SVM classifier with the EMD-based kernel outperformed the baselines and produced ROC AUC value of 0.71. 

This study generally confirms that spectral distributions of the normalized Laplacians capture some meaningful structural properties of brain networks which make them different from other network classes, and also suggests that spectral distributions of connectomes can help to distinguish normal and pathological brain development. Still, further studies are needed to explore whether these findings can generalize to other classification tasks and, importantly, other schemes of network construction (in terms of both definition of nodes and weighting of edges). Also, in this study we only discussed structural connectomes; spectral properties of functional brain networks are to be studied.        




\end{document}